\documentclass[a4paper, 10pt, conference]{ieeeconf}      
\usepackage{FG2019}

\FGfinalcopy 

\IEEEoverridecommandlockouts                              
\usepackage{times}
\usepackage[pagebackref=true,breaklinks=true,linkcolor=blue,urlcolor= blue,citecolor=blue,letterpaper=true,colorlinks,bookmarks=false]{hyperref}
\newcommand{\tht}[2]{\begin{tabular}{@{}#1@{}}#2\end{tabular}}

\usepackage{bm,algorithm,algpseudocode,threeparttable,subfigure,rotating,multirow,array,graphics,textcomp}
\usepackage{setspace}
\usepackage{booktabs} 
\usepackage{amsmath}
\usepackage{amssymb}
\usepackage{graphicx}

\usepackage{caption}
\usepackage[keeplastbox]{flushend}

\let\labelindent\relax
\usepackage{enumitem}
\setitemize{noitemsep,topsep=0pt,parsep=0pt,partopsep=0pt}
\setenumerate{noitemsep,topsep=0pt,parsep=0pt,partopsep=0pt}

\makeatletter
\def\endthebibliography{%
	\def\@noitemerr{\@latex@warning{Empty `thebibliography' environment}}%
	\endlist
}

\def\FGPaperID{40} 

\title{\LARGE \bf
Attribute-Guided Sketch Generation
}

\author{\parbox{16cm}{\centering
    {\large Hao Tang$^1$,  Xinya Chen$^2$, Wei Wang$^3$, Dan Xu$^4$, Jason J. Corso$^5$, Nicu Sebe$^1$, Yan Yan$^6$}\\
    {\normalsize
    $^1$University of Trento \, $^2$Huazhong University of Science and Technology \, $^3$EPFL \, \\
	$^4$University of Oxford \, $^5$University of Michigan \, $^6$Texas State University \\
}}
}

\begin{document}
	
\IEEEoverridecommandlockouts\pubid{\makebox[\columnwidth]{978-1-7281-0089-0/19/\$31.00~\copyright{}2019 IEEE \hfill}
	\hspace{\columnsep}\makebox[\columnwidth]{ }}

\ifFGfinal
\thispagestyle{empty}
\pagestyle{empty}
\else
\author{Anonymous FG 2019 submission\\ Paper ID \FGPaperID \\}
\pagestyle{plain}
\fi
\maketitle

\begin{abstract}
	
	Facial attributes are important since they provide a detailed description and determine the visual appearance of human faces. 
	In this paper, we aim at converting a face image to a sketch while simultaneously generating facial attributes. To this end, we propose a novel Attribute-Guided Sketch Generative Adversarial Network (ASGAN) which is an end-to-end framework and contains two pairs of generators and discriminators, one of which is used to generate faces with attributes while the other one is employed for image-to-sketch translation.
	The two generators form a W-shaped network (W-net) and they are trained jointly with a weight-sharing constraint.
	Additionally, we also propose two novel discriminators, the residual one focusing on attribute generation and the triplex one helping to generate realistic looking sketches. 
	To validate our model, we have created a new large dataset with 8,804 images, named the Attribute Face Photo \& Sketch (AFPS) dataset which is the first dataset containing attributes associated to face sketch images. 
	The experimental results demonstrate that the proposed network (i) generates more photo-realistic faces with sharper facial attributes than baselines and (ii) has good generalization capability on different generative tasks. 
	
\end{abstract}

\section{Introduction}

Recently, there has been a new trend in computer vision to use machines to express the ``creativity'' of art. 
Novel and never-seen-before images can be generated by inverting the convolution process in CNN (``upconvolution'' or ``deconvolution''), which gives such networks the ability to ``dream''~\cite{mordvintsev2015inceptionism} and to generate images. 
To implement these tasks, deep generative models have been usually adopted, and these networks have also been employed for face-to-sketch translation. These deep models can be roughly divided into two categories, namely, the Generative Adversarial Networks (GANs) \cite{goodfellow2014generative} and Variational AutoEncoders (VAEs) \cite{kingma2013auto,larsen2015autoencoding,makhzani2015adversarial}.
In GANs, there are two subnetworks, a generator and a discriminator, which play the two-player minimax game with a value function \cite{goodfellow2014generative}.
The generator acts as a mapping function to convert an input image to the generated image so that it fools the discriminator, which is trained to distinguish the generated images from the input real images. In this work, we rely on GANs as the basis to implement the face-to-sketch translation model.

\begin{figure}[!t] \tiny
	\centering
	\includegraphics[width=1\linewidth]{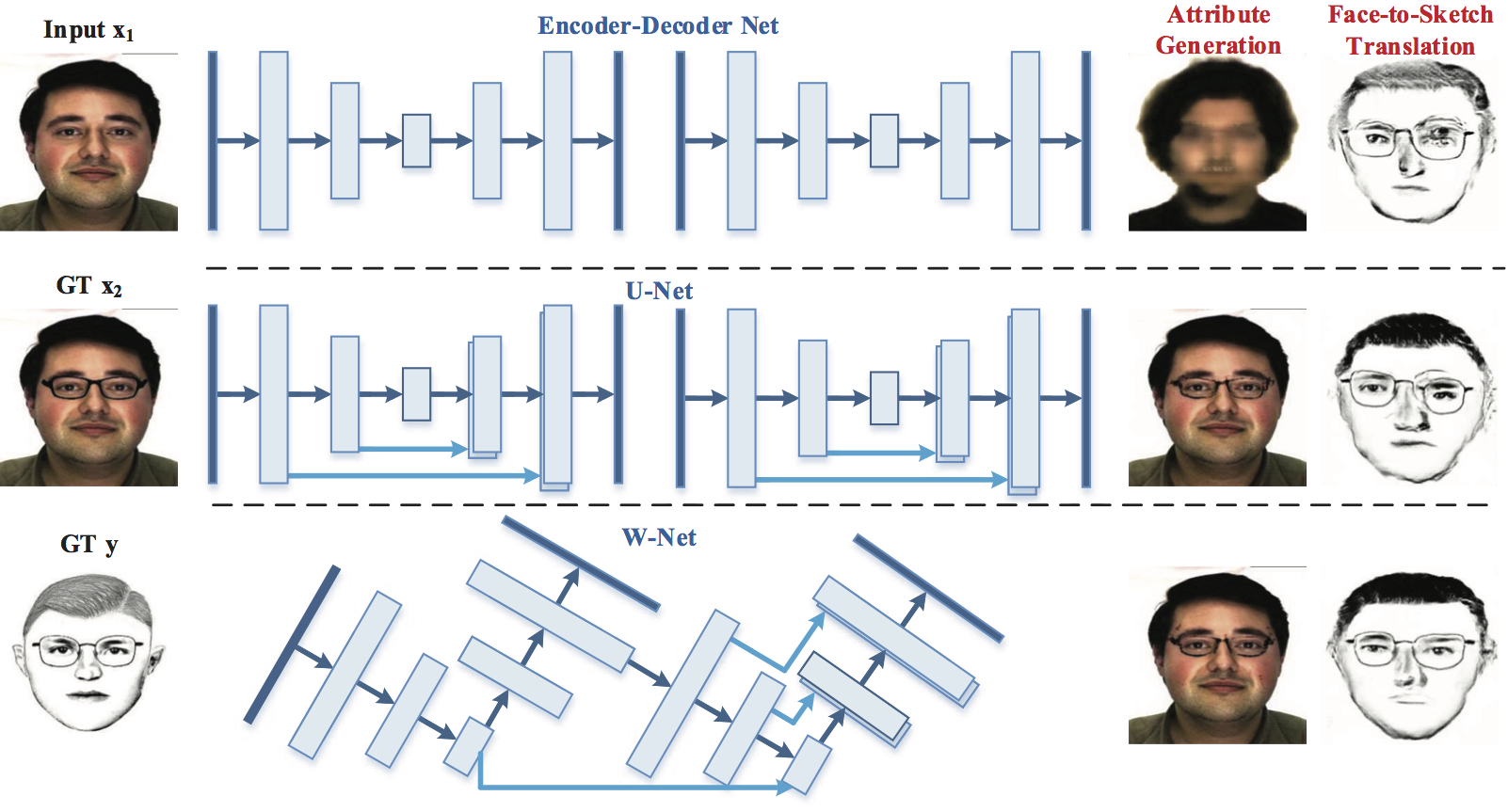}
	\caption{Architecture comparison of three generators. The goal of this work is to generate facial attributes and sketches simultaneously.  To achieve this target, we need to train two Encoder-Decoder networks~\cite{hinton2006reducing} (Top), two U-nets \cite{isola2016image} (Middle) 	or the proposed W-net (Bottom). The W-net consists of two subnetworks which are fused together via a novel joint learning strategy.}
	\label{fig:layout}
	\vspace{-0.6cm}
\end{figure}

Face-to-sketch translation is quite challenging due to the fact that it is a non-linear process conditioned on the appearance of the input face. 
To address this problem, several methods have been proposed \cite{song2017recursive,zhang2017content,zhang2017face,wolf2017unsupervised,lu2018image} for image-to-image translation problems which convert a photo to a sketch. 
However, these works focus only on the face-to-sketch-translation ignoring the possibility of using facial attributes (\emph{e.g.}, expressions, age) and of generating sketches conditioned on external attributes (\emph{e.g.}, glasses, scarf).

\begin{figure*}[!t] \tiny
	\centering
	\includegraphics[width=0.85\linewidth]{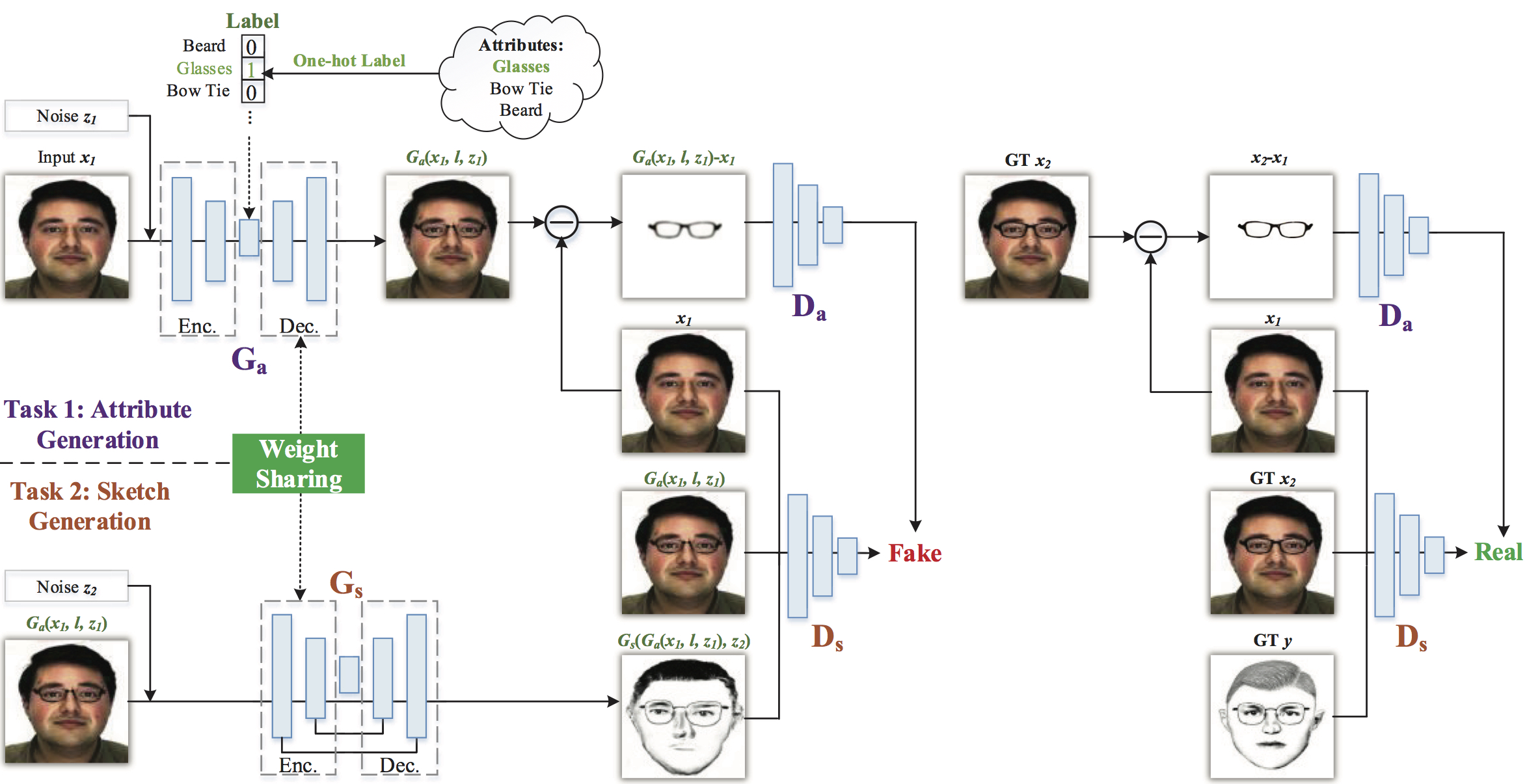}
	\vspace*{0.1cm} 
	\caption{The structure of the proposed ASGAN.
		The generator is a W-net composed of $G_a$ and $G_s$, $G_a$ is an Encoder-Decoder network and $G_s$ is a U-net.
		$G_a$ receives the attribute labels at the bottleneck fully connected layer to guide the attribute generation.
		Both generators are made up of an encoder (Enc.) and a decoder (Dec.).
		We train $G_a$ and $G_s$ in an alternating way
		and they share the same weights between the decoder of $G_a$ and the encoder of $G_s$.
		Moreover, the corresponding discriminators $D_a$ and $D_s$ are the residual and triplex discriminators.}
	\label{fig:framework}
	\vspace{-0.6cm}
\end{figure*}

To overcome this challenging problem, we present a novel Attribute-Guided Sketch Generative Adversarial Network (ASGAN) based on conditional generative adversarial networks.
ASGAN contains two generators and two discriminators. 
The two generators $G_a$ and $G_s$ comprise a novel W-shaped network (W-net) and they are learned jointly as shown in Fig.~\ref{fig:layout}.
We set $G_a$ as an Encoder-Decoder network \cite{hinton2006reducing} and $G_s$ as a U-net \cite{isola2016image}. The proposed W-net can jointly perform facial attribute generation and face-to-sketch translation. 
Our formulation is similar to Pix2pix \cite{isola2016image} but we significantly differ from them by extending it to handle the problem of generating sketches with attributes which can be conditioned not only on the image priors, but also on the attribute labels.
$G_a$ learns the translation from images without attributes to images with attributes guided by attribute labels. In this way, the different attributes can be learned simultaneously as in StarGAN \cite{choi2017stargan} and G$^2$GAN \cite{tang2019dual}. Next, $G_s$ converts images to sketches with the learned attributes. In detail, the first branch in the W-net is a conditional Encoder-Decoder network which learns to add attributes to the input images. This is implemented by fusing the embeddings of the conditioning label and the embeddings of the input face image at the bottleneck. The second branch learns to convert the generated face images with attribute to sketches with attribute. 
We train the two generators $G_a$ and $G_s$ in an alternating way and they share the same weights between the decoder of $G_a$ and the encoder of $G_s$, which limits the network behavior and benefits them from  each other. 
Moreover, two loss functions are designed to train these generators, \emph{i.e.}, the attribute loss and the sketch loss. Besides, compared with the simple combination of two Pix2pix models, the proposed W-net only needs 75\% of parameters by a weight-sharing strategy.

We also propose two novel discriminators $D_a$ and $D_s$, the residual and the triplex discriminator. To focus on learning of facial attributes, the discriminator $D_a$ is trained to distinguish the residuals between the input images and the generated images from the residuals between the input images and the ground truth image which have the conditioned attributes.  
Moreover, since the input image, the attributed image, and the attributed sketch have strong correlations, we take the triplet (face, face with attribute, sketch with attribute) as the input to the triplex discriminator $D_s$ which is used to distinguish the real triplet from the fake one as shown in Fig. \ref{fig:framework}. In this way, $D_s$ could also take into consideration the correlations between the elements. To evaluate the quality of a generated image, we present a novel evaluation metric namely the Feature-Level Similarity Score (FLSS) inspired by feature matching. FLSS works as a complementary metric to Inception Score (IS) and Self Similarity Matrix (SSIM). 
In addition, to validate the proposed network and the overall framework, we have collected a new dataset of 8,804 images collected from five existing datasets by adding visual attributes (glasses, beard, and bow tie), named the Attributed Face Photo \& Sketch (AFPS).  
AFPS consists of two subsets, \emph{i.e.}, face images with attributes and face sketches with attributes. To our knowledge, this is the first dataset containing attributes associated to face sketch images.
Experimental results demonstrate that the proposed ASGAN (i) generates more photo-realistic faces with sharper/more realistic facial attributes than baselines on the  AFPS dataset and (ii) has a good generalization ability on other generative tasks, \emph{i.e.}, face colorization and face completion.
In summary, the contributions of this paper are as follows:
\begin{itemize}[leftmargin=*]
	\item We  propose a novel Attribute Guided Sketch Generative Adversarial Network (ASGAN). The two generators in ASGAN form a novel W-net using a weight-sharing paradigm, which can generate faces with attributes and the corresponding sketches jointly.
	\item We design two novel discriminators, the residual and the triplex discriminators.
	The former one focuses on attribute generation by distinguishing the generated attribute from the real attribute. The latter one focuses on sketch translation by distinguishing the real triplet tuple (original face, real face with attribute, real sketch with attribute) from the fake triplet tuple (original face, generated face with attribute, generated sketch with attribute).
	\item We introduce a new AFSP dataset, which contains 8,804 face images and sketches with visual attributes. The new dataset will be made available to the research community. 
\end{itemize}
The proposed FLSS metric and AFPS dataset are  available at \url{https://github.com/Ha0Tang/ASGAN}.

\section{Related Work}
\label{sec:relatewprk}

Many works have tried to address the face-to-sketch problem. For instance, Song \emph{et al.} \cite{song2017recursive} present the Bidirectional Transformation Network (BTN), which generates a whole face/sketch recursively by using a small number of facial patches. 
Zhang \emph{et al.} \cite{zhang2017face} propose to integrate a Sparse Representation-based Greedy Search (SRGS) and Bayesian Inference (BI) for face sketch synthesis. 
However, all of these works focus on the face-to-sketch-translation task, but they could not generate sketches conditioned on external facial attributes. Generating facial attributes (\emph{e.g.}, glasses, scarf, facial expression, age) associated to the sketch is very challenging.

Recently, Generative Adversarial Networks (GANs) \cite{goodfellow2014generative} have drawn significant attention in both  supervised and unsupervised learning research fields, and a lot of GAN variants have been explored.
For example, the Conditional GANs (CGANs) are introduced to solve ill-posed problems, such as text-to-image translation \cite{reed2016generative,mansimov2015generating}, image-to-image translation \cite{isola2016image}, video-to-video translation \cite{wang2018video}.
CGANs incorporate additional information into GANs, \emph{e.g.}, category labels~\cite{perarnau2016invertible,odena2016conditional,tang2019dual}, 
text description~\cite{zhang2016stackgan,reed2016generative}, object keypoints~\cite{ma2017pose,reed2016learning}, human skeleton \cite{tang2018gesturegan,siarohin2018deformable}, context~\cite{denton2016semi}, segmentation maps \cite{mo2018instagan}
and conditional images~\cite{isola2016image,yoo2016pixel}.

Based on CGANs, Isola \emph{et al.} \cite{isola2016image} have developed a generic framework ``Pix2pix'', which is suitable for different generative tasks.
In Pix2pix, one conditional image is adopted as a reference during the training time.
The generator in Pix2pix is a U-net, which tries to synthesize a fake image conditioned on the given conditional image in order to fool the discriminator, while the discriminator tries to identify the fake image by comparing it with the corresponding target image.
Under these settings, the discriminator takes the pairs of images as input.
The U-net is actually an Encoder-Decoder network with skip connection, in which the encoder consists of multiple convolution layers and the decoder consists of multiple deconvolution layers. Isola \emph{et al.}~\cite{isola2016image} added skip connections between each layer $i$ and layer $n{-}i$ which allows feature sharing between the encoder and decoder, where $n$ is the total number of layers. 
All channels at layer $i$ are simply concatenated with those at layer $n{-}i$ by the skip connections.
\cite{isola2016image} shares a similar goal with us, but it cannot solve the face-to-attributed-sketch translation task since it cannot convert a face image to a sketch conditioned on external facial attributes while our ASGAN is specifically designed to tackle this task.

\section{Model Description}
\label{sec:method}

\subsection{Objective Function}
\label{subsec:objectivefunction}
GANs are generative models that learn the mapping from a random noise $z$ to the output image $y$, $G {:} z {\rightarrow} y$~\cite{goodfellow2014generative}. 
The objective function of a traditional GAN \cite{goodfellow2014generative} can be formulated as follows:
\begin{equation}\small
\begin{aligned}
& \mathcal{L}_{GAN}(G, D) = \\
& \mathbb{E}_{y\sim{p_{\rm data}}(y)}[\log D(y)]
+ \mathbb{E}_{z\sim{P_{z}(z)}}[\log (1 - D(G(z)))].
\end{aligned}
\end{equation}

In contrast, CGANs \cite{isola2016image} learn the mapping from a conditional image $x$ and a random noise $z$ to $y$, $G {:} \{x, z\} {\rightarrow} y$. 
The generator $G$ is trained to produce outputs that cannot be distinguished from ``real'' images by an adversarial discriminator $D$, while the discriminator is trained to distinguish the ``fake'' images from the ``real'' ones. 
The objective function of a CGAN \cite{isola2016image} can be expressed as follows:
\begin{equation}\small
\begin{aligned}
& \mathcal{L}_{cGAN}(G, D) = \mathbb{E}_{x, y\sim{p_{\rm data}}(x, y)}[\log D(x, y)] \\
& + \mathbb{E}_{x\sim{p_{\rm data}}(x), z\sim{p_{z}(z)}}[\log (1 - D(x, G(x, z)))].
\end{aligned}
\end{equation}
where $G$ tries to minimize the objective function while $D$ tries to maximize it.

Our ASGAN contains two generator/discriminator pairs, with
$G_a/D_a$ controls facial attribute generation and and $G_s/D_s$ controls face-to-sketch translation, respectively. 
Similar to StarGAN \cite{choi2017stargan} and G$^2$GAN \cite{tang2019dual}, all the attributes can be learned simultaneously as $G_a$ can receive arbitrary facial attribute label.
The facial attribute is represented by a one hot vector which is used to distinguish each attribute from others. 
In the hot vector, only the element which corresponds to the label is set to 1 while the others are set to 0. 
The one hot vector is passed to a linear layer to get a feature embedding with 64 dimensions and then the embeddings are concatenated with an image embedding vector and passed to the fully connected layer at the bottleneck  of $G_a$. 
This training procedure is shown in Fig.~\ref{fig:framework}.
The loss functions of the facial attribute generation network and face-to-sketch translation network write as,
\begin{equation} \small
\begin{aligned}
& \mathcal{L}_{attribute}(G_a, D_a) = \mathbb{E}_{x_1, x_2\sim{p_{\rm data}}(x_1, x_2)}[\log D_a(x_2-x_1)]\\
&  + \mathbb{E}_{x_1, \widehat{x_2}\sim{p_{\rm data}}(x_1, \widehat{x_2})}[\log (1 - D_a(\widehat{x_2}- x_1))],  
\end{aligned}
\end{equation}
\vspace{-0.4cm}
\begin{equation}\small
\begin{aligned}
& \mathcal{L}_{sketch}(G_s, D_s) = \mathbb{E}_{x_1, x_2, y\sim{p_{\rm data}}(x_1, x_2, y)}[\log D_s(x_1, x_2, y)]  \\
& + \mathbb{E}_{x_1, \widehat{x_2}, \widehat{y}\sim{p_{\rm data}}(x_1, \widehat{x_2}, \widehat{y})}[\log (1 - D_s(x_1, \widehat{x_2}, \widehat{y}))],
\end{aligned}
\end{equation}
where $x_1$ represents the input image without attribute; $x_2$ denotes the ground truth image with attribute, and $y$ is the ground truth sketch with attribute.
The two noises $z_1$ and $z_2$ are sampled independently.
The outputs are not two sketches but one face image with attribute $\widehat{x_2}=G_a(x_1, l, z_1)$ and one sketch with attribute $\widehat{y}=G_s(G_a(x_1, l, z_1), z_2)$.

Previous approaches of CGANs have found it beneficial to mix the GAN objective function with a more traditional loss, such as the $L1$ \cite{isola2016image} or $L2$ losses \cite{pathak2016context}. 
Under that condition, the generator is asked to not only fool the discriminator but also to be closer to the ground truth output in a $L1$ or $L2$ sense.
We also explore this option, using $L1$ distance rather than the $L2$ since $L1$ encourages less blurring~\cite{isola2016image}:
\begin{equation}\small
\mathcal{L}_{L1}(G_a) = \mathbb{E}_{{x_2, \widehat{x_2}\sim{p_{\rm data}}(x_2, \widehat{x_2})}}[||x_2 - \widehat{x_2}||_1].
\end{equation}
\vspace{-0.4cm}
\begin{equation}\small
\mathcal{L}_{L1}(G_s) = \mathbb{E}_{{y,\widehat{y}\sim{p_{\rm data}}(y, \widehat{y})}}[||y - \widehat{y}||_1].
\end{equation}
It is worth noting that there are many loss functions we could try, \emph{e.g.}, feature loss \cite{johnson2016perceptual} and total variation loss \cite{mahendran2015understanding}.
We did not use them in our experiments and we  list them here for completeness.
Therefore, the final objective is:
\begin{equation}\small
\begin{aligned}
G^\ast = & \arg \mathop{\min}\limits_{G_a,G_s} \mathop{\max}\limits_{D_a,D_s} \mathcal{L}_{attribute}(G_a, D_a) + \\ 
& \mathcal{L}_{sketch}(G_s, D_s) + \lambda \mathcal{L}_{L1}(G),
\end{aligned}
\label{equ:loss}
\end{equation} 
where $\mathcal{L}_{L1}(G)=\mathcal{L}_{L1}(G_a)+\mathcal{L}_{L1}(G_s)$.
In our experiments, instead of using Gaussian noises $z_1$ and $z_2$, we generate the noise in the dropout layer, which is consistent with \cite{isola2016image}.

\vspace{-0.1cm}
\subsection{Network Architectures}
\label{subsec:networkarchitectures}
The architectures of generators and discriminators in this paper are developed based on \cite{isola2016image}.
The face-to-sketch translation problem is essentially the problem that maps a high resolution input face image to a high resolution output sketch image. 
In addition, although the input image and the output sketch differ in appearance, the sketch is a rendering of the real image and they share the same contour. 
Therefore, the contour of the input image is roughly aligned with the contour of the output sketch. 
Many previous solutions \cite{pathak2016context,wang2016generative,johnson2016perceptual} to this problem usually employ an Encoder-Decoder network~\cite{hinton2006reducing}. 
In such a network, the input is passed through a series of layers and is progressively down-sampled, until a bottleneck layer is reached, at which point the process is reversed. 
Such a network requires that all information flows through all the layers, including the bottleneck. 
For many image translation problems, there is a great deal of low-level information shared between the input and output, and it would be desirable to transfer this information directly across the net.
For example, in the case of face-to-sketch translation, the input image and the output sketch share the location of prominent structures and edges.
Therefore, the design of the generator architecture in \cite{isola2016image} takes this into consideration.
To help the generator circumvent the bottleneck for information like this, \cite{isola2016image} adds skip connections, following the general shape of a U-Net \cite{ronneberger2015u}.

\noindent\textbf{Generator Architectures.}
In this paper, we present a novel generator which is a W-shaped network consisting of two generators $G_a$ and $G_s$.
The components of the W-shaped network are shown in the following. 

\textbf{Generator $G_a$:}
The encoder in $G_a$ has eight Convolution-BatchNorm-ReLu layers, while the decoder also has eight layers, \emph{i.e.}, three Convolution-BatchNorm-Dropout-ReLu layers and five Convolution-BatchNorm-ReLu layers.
The numbers of convolution layer feature maps in the encoder are $[64, 128, 256, 512, 512, 512, 512, 512]$.
The numbers of convolution layer feature maps in the decoder are $[512, 512, 512, 512, 512, 256, 128, 64]$.
In the encoder and decoder, the kernels in all the convolution layers have the same shape which is $4{\times}4$ and the same stride which is 2. 
The output of each convolution layer in the encoder is down-sampled by 2 while the output of each convolution layer in the decoder is up-sampled by 2.
At the end of the decoder, a convolution is applied to map it to a 3 channels result, followed by a tanh function.
BatchNorm is not employed into the first layer in the encoder.

\textbf{Generator $G_s$:}
We aim to generate the attributed face and the corresponding sketch jointly. Therefore, we select U-net as the sketch generation model as it can reuse the features of the attributed face via the skip connection. 
In this way, we formulate the problem as a multi-task learning problem and both attributed face and the corresponding sketches could benefit from each other.
Generator $G_s$ is a U-net, which is an Encoder-Decoder net with skip connections between mirrored layers in the encoder and decoder stacks~\cite{ronneberger2015u}.
The number of the convolution layer feature maps of the encoder in $G_s$ is the same with the decoder in $G_a$. 
Besides, the convolution layers in the encoder of $G_s$ share the weights with the decoder in $G_a$, which makes the final output not only learn the facial attribute but also the face-to-sketch translation.
However, the skip connections change the number of channels in the decoder, thus the convolution layer feature map numbers in the decoder of the generative net are $[512, 1024, 1024, 1024, 1024, 512, 256, 128]$.

\noindent\textbf{Discriminator Architectures.}
For $D_a$ and $D_s$ we adopt the same architectures in \cite{isola2016image}, which are built with the basic Convolution-BatchNorm-ReLU layer. 
All ReLUs are leaky, with slope 0.2.
The convolution layer feature map numbers in this discriminator are $[64, 128, 256, 512, 512, 512]$.
After the last layer, a convolution is applied to map it to a 1-D value, followed by a sigmoid function.
BatchNorm is not adopted into the first layer.

\subsection{Optimization}

We follow the standard optimization method from \cite{goodfellow2014generative} to optimize the proposed ASGAN, \emph{i.e.}, we alternate between one gradient descent step on discriminator $D_a$ and $D_s$ with $G_a$ and $G_s$ fixed, then one step on generator $G_a$ and $G_s$ with $D_a$ and $D_s$ fixed.
In addition, following the suggestion made in \cite{goodfellow2014generative}, we train the generator $G_a$ (or $G_s$) to maximize $D_a(\widehat{x_2}- x_1)$ (or $D_s(x_1, \widehat{x_2}, \widehat{y})$) rather than minimizing $\log (1 - D_a(\widehat{x_2}- x_1))$ (or $\log (1 - D_s(x_1, \widehat{x_2}, \widehat{y}))$).
Moreover, in order to slow down the rate of the discriminator $D_a$ (or $D_s$) relative to the generator $G_a$ (or $G_s$) we divide the objective by 2 while optimizing $D_a$ (or $D_s$).
We employ the Adam optimizer \cite{kingma2014adam} as solver, the momentum terms $\beta_1$ and  $\beta_2$ of Adam are 0.5 and 0.999, respectively.  
The initial learning rate for Adam is 0.0002.

\subsection{Feature-Level Similarity Score}
\label{subsec:evaluationmetric}

Currently, there are two kinds of evaluation metrics for the image generation task, \emph{i.e.}, qualitative and quantitative.
On one hand, the qualitative results are usually studied by human observers and machines, \emph{e.g.}, previous work~\cite{siarohin2018deformable} conduct a user survey on generated images and collected the scores given by the users. 
Others~\cite{isola2016image,zhang2016colorful} ran ``real vs fake'' perceptual studies on Amazon Mechanical Turk (AMT).
On the other hand, the quantitative results are usually evaluated by algorithms, \emph{e.g.}, \cite{perarnau2016invertible,deshpande2016learning,guccluturk2016convolutional,yan2016attribute2image} take an image distance, such as, MSE (Mean Square Error), SSIM (Structural Similarity) or PSNG (Peak Signal to Noise Ratio) to measure the quality of the results.
Others apply the Inception or ResNet models to measure the generated images \cite{szegedy2016rethinking,odena2016conditional,tang2018gesturegan}.
Recent works adopt additional classifiers to predict if the generated images can be correctly detected \cite{wang2016generative}, segmented \cite{isola2016image} or classified \cite{iizuka2016let,salimans2016improved}.

It is clear that there is no single metric that can be used to measure the quality of the generated images accurately and holistically \cite{salimans2016improved}.
However, measurements such as SSIM and IS (Inception Score) are not good metrics as shown in \cite{shi2016real,johnson2016perceptual,ma2017pose} since sometimes the generated images with more sharper and more photo-realistic aspects have a lower SSIM and IS.
We, therefore, present an alternative method to measure the quality of generated images called Feature-Level Similarity Score (FLSS), which is similar to the pixel-level similarity method \cite{yoo2016pixel}.
The idea of the FLSS derives from the feature matching methods in image retrieval~\cite{tangnovel}.
Let us define the generated images $I_g^i$ and the ground truth images $I_t^i$ (where $i$ denotes the index of generated or ground truth images).
Given a function of feature extraction $F$, our task is to extract features from $I_g^i$ and $I_t^i$ and match them: 
$F(I_g^i) \bigcap F(I_t^i)$, where $\bigcap$ denotes the number of matches between two sets of descriptors $F(I_g^i)$ and $F(I_t^i)$.
In other words, we do feature matching between two sets of descriptors, then calculate how many features (matches) are shared between the two sets:
\begin{equation} \small
S_\text{FLSS} = \frac{1}{N} \sum_{i = 1}^N \frac{\phi(F(I_g^i) \bigcap F(I_t^i))}{\min(\phi(F(I_g^i)), \phi(F(I_t^i)))},
\label{equ:score}
\end{equation} 
where $\phi(\cdot)$ defines the cardinality of the set $(\cdot)$, $N$ denotes the number of the generated or ground truth images in a dataset and $S_\text{FLSS}$ is the final similarity score of the dataset.

\begin{table}[t!] \small
	\centering
	\caption{Key characteristics of the proposed AFPS dataset.}
	\label{tab:dataset}
	\resizebox{1\linewidth}{!}{%
		\begin{tabular}{l|c|c|c|c|r}
			\toprule
			Dataset         			                      & Image Source                & Type        & Training & Testing & Total\\
			\midrule
			CUFS \cite{wang2009face}              & CUHK student                     & Hand-Drawn  & 94             & 94            & 188 \\ \hline
			CUFSF \cite{zhang2011coupled}         & FERET \cite{phillips2000feret}   & Hand-Drawn   & 562            & 561           & 1,123 \\ \hline	
			E-PRIP \cite{mittal2014recognizing}   & AR \cite{martinez1998ar}         & Composite   & 70             & 16            & 86 \\ \hline  
			PRIP-VSGC \cite{klum2014facesketchid} & AR \cite{martinez1998ar}         & Composite  & 70             & 16            & 86 \\ \hline
			Caricature \cite{klare2012towards}    & Internet                         & Caricatural & 101            & 30            & 131 \\ \hline
			Total                                 & AFPS dataset                     &   -         & 897            & 717           & 1,614  \\
			\bottomrule
	\end{tabular}}
	\vspace{-0.6cm}
\end{table}

\begin{table*}[!t] \tiny
	\centering
	\caption{Quantitative results of the generated sketches using different losses on the proposed APFS dataset, compared with Encoder-Decoder \cite{hinton2006reducing} and Pix2pix \cite{isola2016image}. For SSIM and FLSS, higher is better. For the table cell of Caricature, the top value is the result of the colorization (col.) task and the bottom one is the result of the completion  (com.)  task.}
	\resizebox{1\linewidth}{!}{%
		\begin{tabular}{lc|cc|cc|cc|cc|cc}
			\toprule
			& & \multicolumn{2}{c|}{CUFS}                   & \multicolumn{2}{c|}{CUFSF} & \multicolumn{2}{c|}{E-PRIP} & \multicolumn{2}{c|}{PRIP-VSGC} &  \multicolumn{2}{c}{Caricature} \\ \cline{3-12}
			Model                                           & Task                & SSIM  & FLSS  & SSIM  & FLSS  & SSIM  & FLSS   & SSIM  & FLSS  & SSIM                    & FLSS \\ \midrule
			attribute loss + sketch loss                    & \tht{c}{col.\\com.} & 0.5090& 0.1571& 0.3454& 0.1225& 0.3761& 0.0790 & 0.3373&0.0781 & \tht{c}{0.2466\\0.8333} & \tht{c}{0.4554\\0.6172}\\ \hline
			\tht{l}{attribute + sketch + $\lambda L1$\\ ASGAN (Ours)}& \tht{c}{col.\\com.} & \textbf{0.6249}& \textbf{0.2534}& \textbf{0.3515}& \textbf{0.1567}& \textbf{0.4864}& \textbf{0.1280} & \textbf{0.4122}&\textbf{0.1258} & \tht{c}{\textbf{0.5830}\\\textbf{0.9109}} & \tht{c}{\textbf{0.5686}\\\textbf{0.6774}}\\ \hline
			Encoder-Decoder \cite{hinton2006reducing}       & \tht{c}{col.\\com.} & 0.5025& 0.1387& 0.3440& 0.1386& 0.4677& 0.1190 & 0.4030&0.1119 & \tht{c}{0.3719\\0.3910} & \tht{c}{0.0391\\0.0491}\\ \hline	
			Pix2pix \cite{isola2016image}                   & \tht{c}{col.\\com.} & 0.6127& 0.2431& 0.3488& 0.1517& 0.4781& 0.1264 & 0.4018&0.1136 & \tht{c}{0.5477\\0.6842} & \tht{c}{0.4724\\0.4100}\\ \hline
			Time (s)  					                    &                     & 239.9 & 15.3  & 1416.8& 102.6 & 40.3  & 2.7    & 40.5  &2.5    & 85.8                    & 5.9     \\ 
			\bottomrule
	\end{tabular}}
	\label{tab:evaluation_all}
	\vspace{-0.3cm}
\end{table*}

\begin{table*}[t!] \footnotesize
	\centering
	\caption{AMT ``real vs fake'' test of the generated attributed faces (attr.) and attributed sketches (sket.) on the APFS dataset, compared with Encoder-Decoder \cite{hinton2006reducing} and Pix2pix \cite{isola2016image}.}
	\begin{tabular}{l|cc|cc|cc|cc}
		\toprule
		\% Tukers label \textit{real}                          & \multicolumn{2}{c|}{CUFS}  & \multicolumn{2}{c|}{CUFSF} & \multicolumn{2}{c|}{E-PRIP} & \multicolumn{2}{c}{PRIP-VSGC} \\ \midrule
		Model                                                  & attr.  & sket.  & attr.  & sket.   & attr.  & sket.  & attr.  & sket. \\ \midrule
		Encoder-Decoder \cite{hinton2006reducing}              & 1.3\%  & 0.7\%  & 1.1\%  & 15.3\%  & 3.1\%  & 0.3\%  & 2.8\%  & 1.6\% \\ \hline
		Pix2pix \cite{isola2016image}                          & 40.1\% & 3.3\%  & 29.3\% & 10.7\%  & 40.7\% & 3.2\%  & 11.8\% & 3.2\% \\ \hline
		ASGAN (Ours)                                           & \textbf{42.7\%} & \textbf{6.9\%}  & \textbf{46.7\%} & \textbf{20.4\%}  & \textbf{43.7\%} & \textbf{4.6\%}  & \textbf{15.6\%} & \textbf{6.3\%} \\ \bottomrule
	\end{tabular}
	\label{tab:evaluation_amt}
	\vspace{-0.4cm}
\end{table*}

\vspace{-0.3cm}
\section{Experiments}
\label{sec:experiment}

In this section, we first introduce the dataset used and the implementation details. 
We then show detailed qualitative and quantitative results and analyses.

\subsection{Experimental Setup}
\noindent \textbf{Datasets.}
To evaluate the proposed method and the overall framework, we have investigated the available face sketch datasets \cite{peng2016face}. 
However, none of them considers attributes on sketches.
Therefore, we use a public photo-editing software to produce the attributes (glasses, beard and bow tie) based on five existing datasets, which are then incorporated into the Attributed Face Photo \& Sketch (AFPS) dataset.
The AFPS contains three types of face sketches, \emph{i.e.}, hand-drawn, composite, and caricatural.
The main features of the AFPS are listed in Table~\ref{tab:dataset}. 
The AFPS contains 897 training samples and 717 testing samples.
Each sample has two images, \emph{i.e.}, a face photo with attribute and a face sketch with attribute.
Since several face images in Caricature, R-PRIP and PRIP-VSGC datasets do not have the room to allow augmenting them with a bow tie, and there are already faces with mustache in the Caricature dataset, we remove the bow tie and beard attributes in Caricature dataset, and the bow tie attribute in both R-PRIP and PRIP-VSGC datasets.
Therefore the total number of images in AFPS is 8,804.

\noindent \textbf{Implementation Details.}
The proposed network is implemented under the deep learning framework using PyTorch.
In our experiments, all the images are re-scaled to $256{\times}256{\times}3$ and all models of each dataset were trained for 200 epochs.
For all datasets, we do left-right flip for data augmentation.
During training time, we need to input $x_1$, $x_2$ and $y$.
While during testing time we only require $x_1$, and the model will output $G_a(x_1, l, z_1)$ and $G_s(G_a(x_1, l, z_1), z_2)$.
The weights are initialized from a Gaussian distribution with mean 0 and standard deviation 0.02.
Training and testing stages are conducted on a Nvidia TITAN Xp GPU with 12 GB memory. 

\begin{figure}[!t] \tiny
	\centering
	\includegraphics[width=\linewidth]{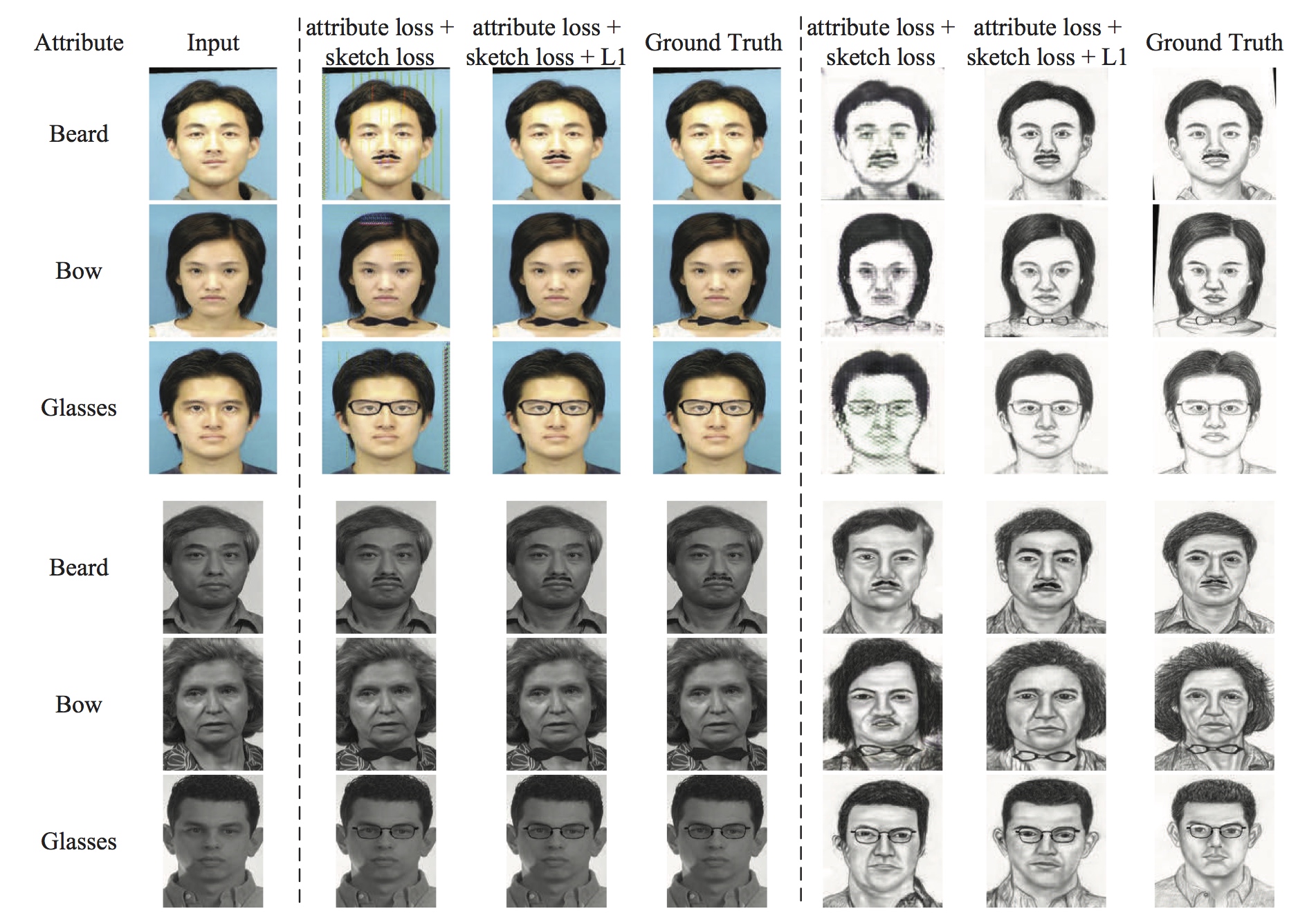}
	\caption{Results of the proposed ASGAN for face-to-attributed-sketch task with different losses on the CUFS (Top) and the CUFSF (Bottom) datasets. Ground truths are provided only for comparison purpose.}
	\label{fig:result_hand_drawn_style}
	\vspace{-0.4cm}
\end{figure}

\subsection{Experimental Results}

\begin{figure}[!t] \tiny
	\centering
	\includegraphics[width=1\linewidth]{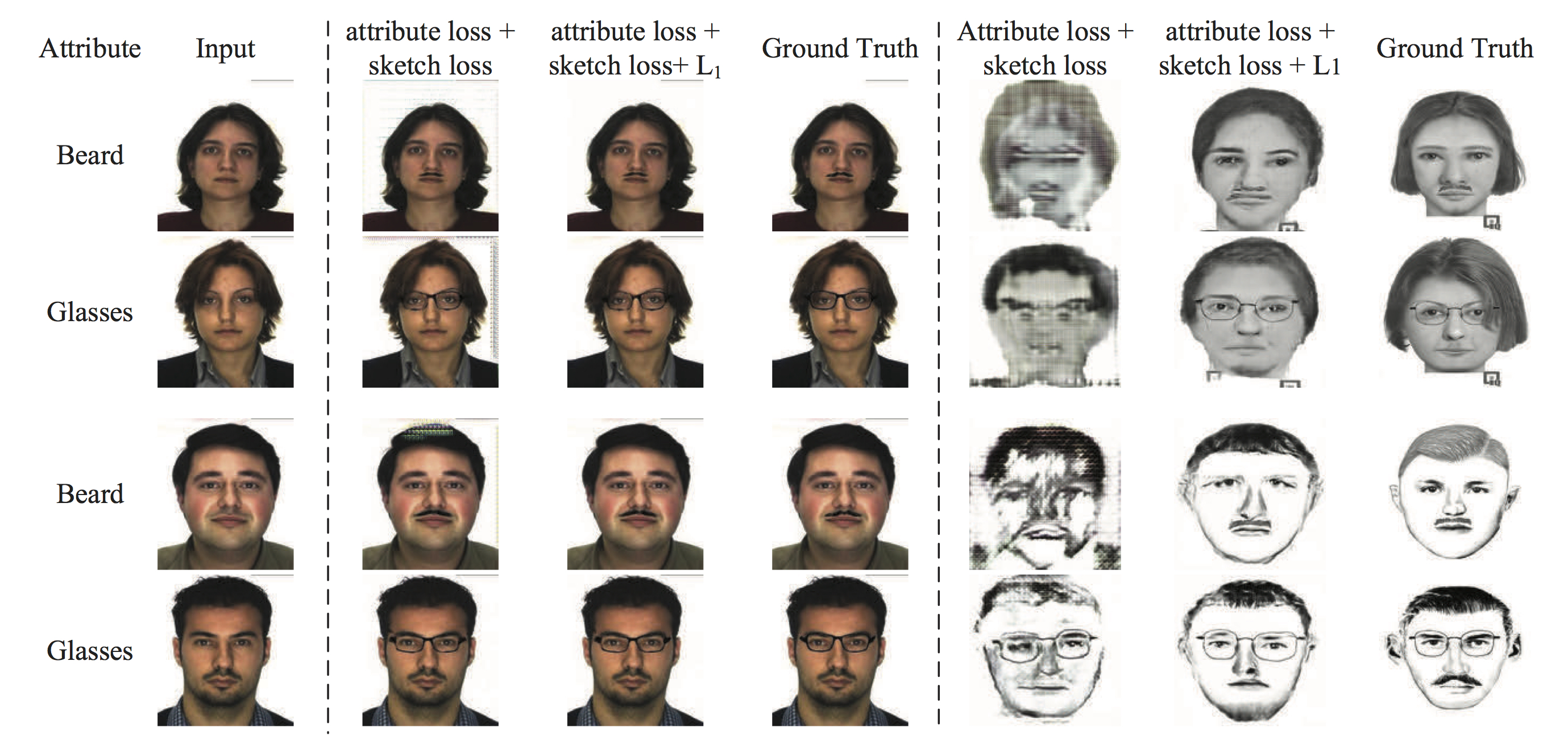}
	\caption{Results of the proposed ASGAN on the face-to-attributed-sketch translation task on the E-PRIP (Top) and PRIP-VSGC (Bottom) datasets. Ground truths are provided only for comparison purpose.}
	\label{fig:result_eprip_and_pripvsgc}
	\vspace{-0.8cm}
\end{figure}

\noindent\textbf{Analysis of Loss Function.}
We first investigate the performance of the proposed ASGAN using the objective in Eq.~\ref{equ:loss}. 
Ablation experiments are run to isolate the effect of the attribute loss term, the sketch loss term and the $L1$ loss term. 
Fig.~\ref{fig:result_hand_drawn_style} and \ref{fig:result_eprip_and_pripvsgc} show the qualitative results of the variations of face-to-attributed-sketch translation task.
The attribute term gives faces attributes.
The sketch term controls face-to-sketch translation.
Finally, the $L1$ term with attribute and sketch terms together can generate photo-realistic sketches and sharp facial attributes simultaneously.
Note that the ground truth sketches on the E-PRIP and PRIP-VSGC datasets do not have torsos such that we do not show the generated results with bow tie attribute.
We also try to generate a beard on the face of a lady as shown in the first row of Fig.~\ref{fig:result_eprip_and_pripvsgc}, which only to validate that the proposed ASGAN is able to generate the correct attributes. 
At testing stage, we are able to control the proposed ASGAN to generate the desired attributes.

We also provide the quantitative results of the generated sketches.
SSIM \cite{wang2004image} and the proposed FLSS are adopted to measure the quality of the generated images. 
The SIFT matching function of the VLFeat library
is applied for the implementation of the proposed FLSS.
We list the average results of all attributes with SSIM and FLSS in Table~\ref{tab:evaluation_all}.
As we can see from Table \ref{tab:evaluation_all}, the attribute + sketch + $\lambda$ $L1$ model produces high-quality results and photo-realistic attributes as compared to other loss combinations.
Beside, we also report the time for evaluating the quality of the generated images.
The proposed FLSS is much faster than SSIM, which highlights the advantage of the proposed evaluation method.

\begin{figure}[!t] \tiny
	\centering
	\includegraphics[width=1\linewidth]{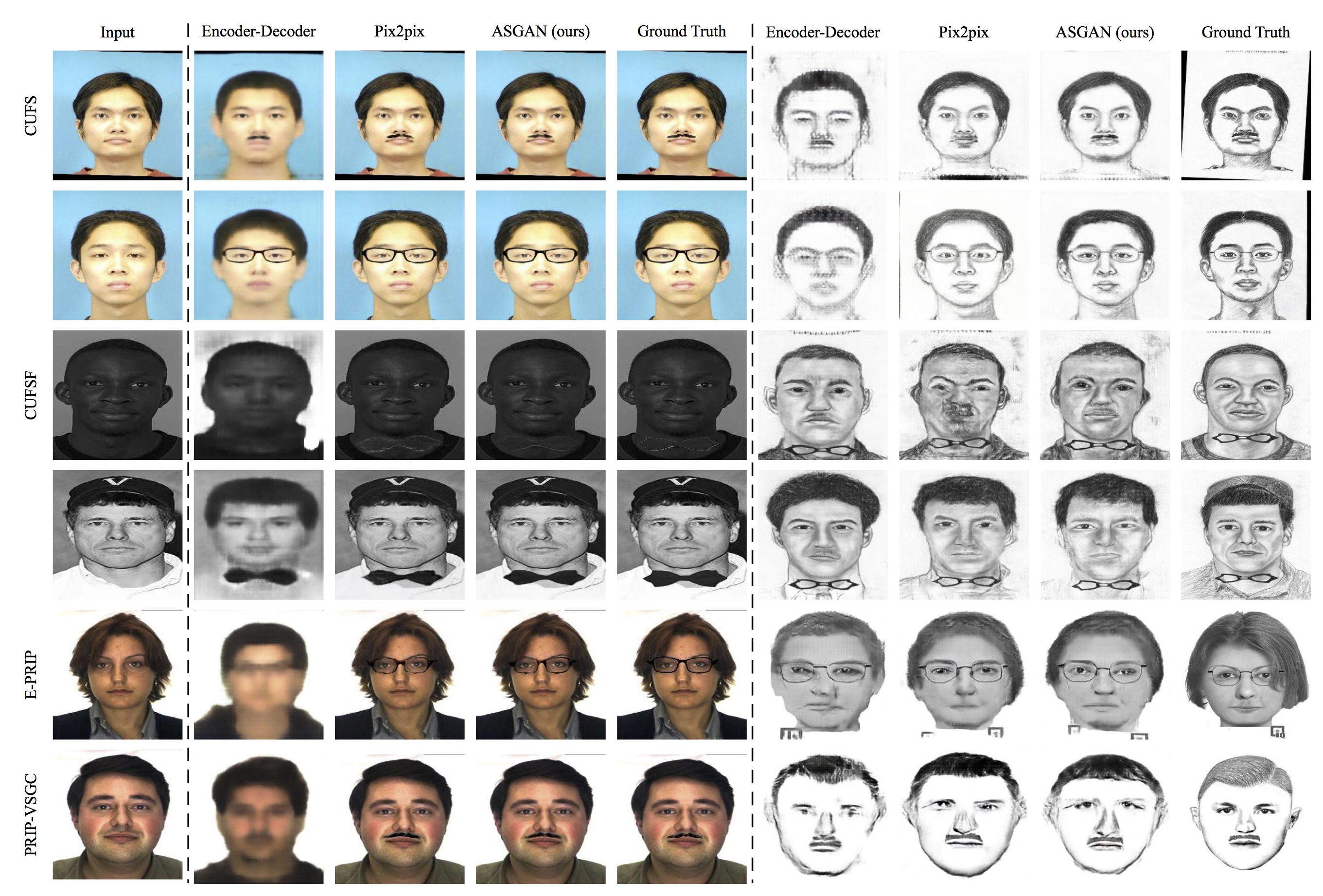}
	\caption{Comparisons of the Encoder-Decoder network~\cite{hinton2006reducing}, Pix2pix~\cite{isola2016image}, ASGAN (ours) and Ground Truths (GT) for face-to-attributed-sketch translation task.}
	\label{fig:result_comparasion}
	\vspace{-0.4cm}
\end{figure}

\noindent \textbf{Comparison against Baselines.}
Our joint learning task tries to generate attributed faces and sketches simultaneously from a single face image, which is quite novel. 
Therefore we only compare our method with the most related ones, \emph{i.e.},  Encoder-Decoder net \cite{hinton2006reducing} and Pix2pix \cite{isola2016image}, which are the most successful models on image-to-image translation.
Note that we need to train two Encoder-Decoder nets or two U-nets, which doubles the parameters, but the proposed ASGAN only has 75\% amount of parameters compared with the baseline methods since we share the parameters between two generators. 
Moreover, for the baselines we need to train these models multiple times for every attribute, while for the proposed ASGAN we only need to train it once.
Results are shown in Fig.~\ref{fig:result_comparasion}  and Table~\ref{tab:evaluation_all}.
We observe that the proposed method achieves superior results compared with Encoder-Decoder \cite{hinton2006reducing} and Pix2pix \cite{isola2016image} in Fig.~\ref{fig:result_comparasion}.
The proposed ASGAN consistently generates clear and convincing visual attributes, and produces more vivid and high-quality face sketches than the baselines.
As we can see in Table \ref{tab:evaluation_all}, the proposed ASGAN consistently outperforms Encoder-Decoder network~\cite{hinton2006reducing} and Pix2pix~\cite{isola2016image} across both metrics on the five datasets.
Moreover, we follow the same protocol from Pix2pix \cite{isola2016image} to run ``real vs fake'' perceptual studies on Amazon Mechanical Turk (AMT).
The average results of all facial attributes are listed in Table \ref{tab:evaluation_amt}.
We can  observe that the proposed ASGAN achieves better performance than Encoder-Decoder net~\cite{hinton2006reducing} and Pix2pix \cite{isola2016image} on both the generated faces and sketches.

\begin{figure}[!tbp] \tiny
	\centering
	\includegraphics[width=1\linewidth]{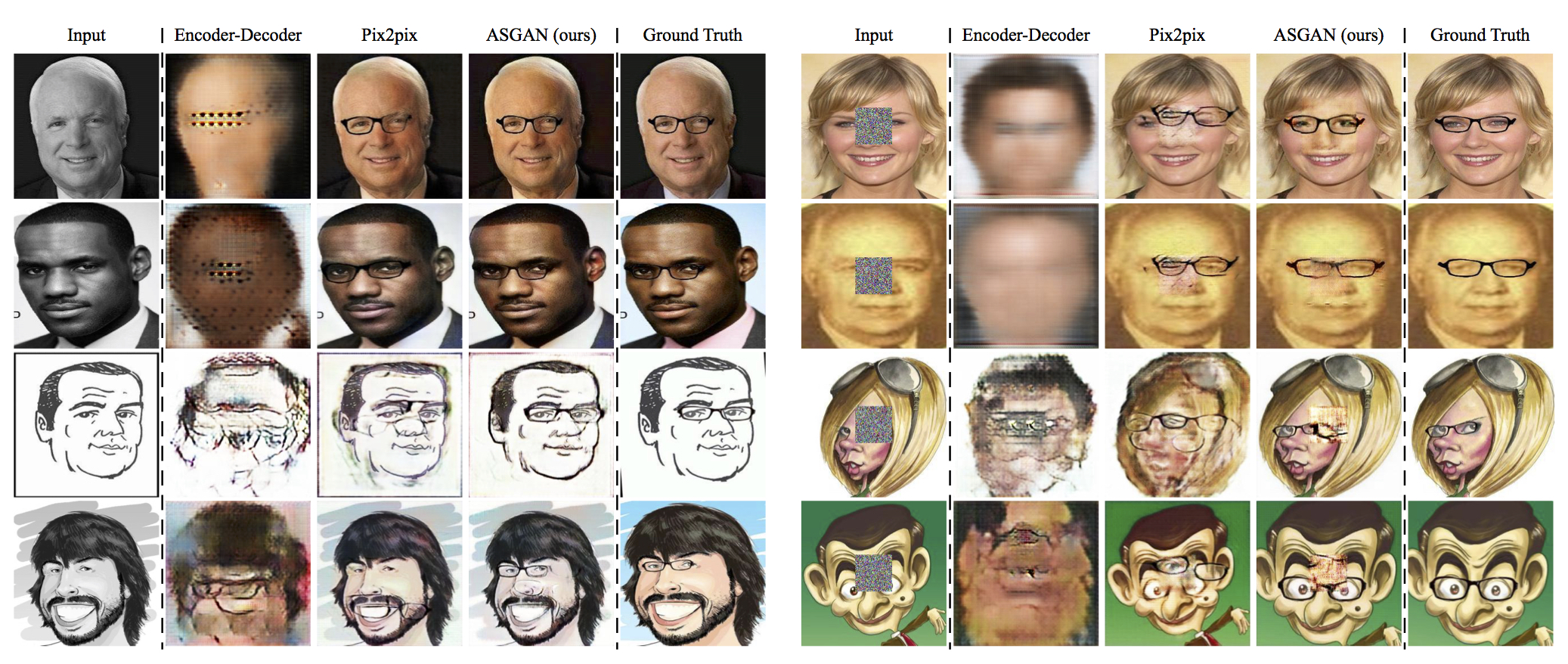}
	\caption{Test results of the Caricature dataset \cite{klare2012towards} on the face colorization (Left) and completion (Right) tasks, compared to Encoder-Decoder network \cite{hinton2006reducing}, Pix2pix \cite{isola2016image}, ASGAN and GT. The input and output images with and without glasses, respectively. For the face completion task, we add random noise in the central square patch in the input image. }
	\label{fig:result_application}
	\vspace{-0.4cm}
\end{figure}

\noindent \textbf{Other Applications.} In addition, to evaluate the effectiveness of the proposed network on different generative tasks, the Caricature dataset \cite{klare2012towards} is employed for face colorization and face completion tasks.
The test results are shown in Fig.~\ref{fig:result_application} and Table~\ref{tab:evaluation_all}.
For face colorization, our approach is quite effective at creating reasonable and realistic color renderings and at the same time generating the right facial attribute.  
For the completion task, the Encoder-Decoder network~\cite{hinton2006reducing} cannot generate face images, while Pix2pix~\cite{isola2016image} is able to generate face images but without convincing attributes. 
We observe that the proposed ASGAN has better qualitative results than both baselines and the missing part can be restored correctly with glasses. 
We obtain good results on both tasks, which indicates that the proposed approach can be useful for other generative tasks.

\section{Conclusion}
\label{sec:conclusion}

In this paper, we present a novel Attributed-Sketch Generative Adversarial Network (ASGAN) which consists two generators and two discriminators.
The generators are jointly trained through a weight-sharing strategy, which can generate attributed face and attributed sketches at the same time.  
We also propose two novel discriminators, the residual and the triplex discriminators.
Experimental results demonstrate that the proposed ASGAN generates more photo-realistic faces with sharper facial attributes than baselines on three different generative tasks, \emph{i.e.}, face-to-attributed-sketch translation, face colorization and face completion.
Finally, the proposed ASGAN has a huge potential for forensic sketch synthesis, which is our future research direction. \\
\noindent \textbf{Acknowledgments.} We acknowledge the National Institute of Standards and Technology Grant 60NANB17D191 for funding this research. We also acknowledge the gift donation from Cisco, Inc for this research.

\footnotesize
\bibliographystyle{ieee}
\bibliography{references}

\end{document}